\documentclass[11pt]{article}
\usepackage{coling2020}
\usepackage{times}
\usepackage{url}
\usepackage{latexsym}

\usepackage{graphicx}
\usepackage{hyperref}
\usepackage{multirow}
\usepackage{color}
\usepackage{xcolor} % From previous preprint
\usepackage{amsfonts,amssymb}
\usepackage{amsmath}
\usepackage{subcaption}
\usepackage{wrapfig}
\usepackage[flushleft]{threeparttable}
\usepackage{placeins}

\colingfinalcopy % Uncomment this line for the final submission

\title{The Impact of Cross-Lingual Adjustment of Contextual Word Representations on Zero-Shot Transfer}

\author{Pavel Efimov\textsuperscript{1}$^{\star}$, Leonid Boytsov\textsuperscript{2}$^{\star\dag}$, Elena Arslanova\textsuperscript{3}, Pavel Braslavski\textsuperscript{3,4} \\
  \textsuperscript{1}ITMO University, Saint Petersburg, Russia\\
    \textsuperscript{2}Bosch Center for Artificial Intelligence, Pittsburgh, USA\\
    \textsuperscript{3}Ural Federal University, Yekaterinburg, Russia\\
    \textsuperscript{4}HSE University, Moscow, Russia\\
  {\tt pavel.vl.efimov@gmail.com, leo@boytsov.info} \\
  {\tt contilen@gmail.com, pbras@yandex.ru} \\}

\date{}

\begin{document}
\def\thefootnote{$\star$}\footnotetext{Contributed equally to the paper.}
\def\thefootnote{$\dag$}\footnotetext{Work done before joining Amazon.}\def\thefootnote{\arabic{footnote}}

\maketitle
\begin{abstract}
  Large multilingual language models such as mBERT or \mbox{XLM-R} enable zero-shot \emph{cross-lingual} transfer in various IR and NLP tasks. 
\newcite{cao2020multilingual} proposed a \emph{data-} and \emph{compute-efficient} method 
for cross-lingual adjustment of mBERT that uses a \emph{small} parallel corpus to make embeddings of related words across languages similar to each other.
They showed it to be effective in NLI for five European languages. 
In contrast we experiment with a typologically diverse set of languages (Spanish, Russian, Vietnamese, and Hindi)
and extend their original implementations to new tasks (XSR, NER, and QA)
and an additional training regime (continual learning).
Our study reproduced gains in NLI for four languages, showed improved NER, XSR, 
and \emph{cross-lingual} QA results in three languages (though some cross-lingual QA gains were not statistically significant), 
while \emph{mono-lingual} QA performance never improved and sometimes degraded. 
Analysis of distances between contextualized embeddings of related and unrelated words (across languages) 
showed that fine-tuning leads to ``forgetting'' some of the cross-lingual alignment information.
Based on this observation, we further improved NLI performance using continual learning.
Our software is publicly available \url{https://github.com/pefimov/cross-lingual-adjustment}.

\end{abstract}

\section{Introduction}

Large \emph{multi-lingual} language models such as mBERT or \mbox{XLM-R}
pre-trained on  large \emph{unpaired} multilingual corpora 
enable zero-shot \emph{cross-lingual}  transfer~\cite{libovicky2019language,pires2019multilingual},
which is sometimes effective even for languages not seen at the pre-training stage~\cite{ebrahimi2021americasnli,muller-etal-2021-unseen}. Contextualized word representations produced by the models
can be further aligned (across languages) with a modest amount of parallel data, e.g., 
by finding a rotation matrix using a bilingual dictionary \cite{mikolov2013exploiting}.
Such a post hoc alignment can improve zero-shot transfer for text parsing and understanding~\cite{kulshreshtha2020cross,wang2019cross,wang2019parsing}. 
This approach is a more data- and computationally-efficient alternative 
to training a machine translation system  
on a large parallel corpus.

\newcite{cao2020multilingual} used a small parallel corpus for direct cross-lingual adjustment of the mBERT model  and found this adjustment procedure to be more effective than the post hoc rotation.
However, they experimented only with a single task: cross-lingual NLI,
and restricted their study to European languages.
In that, we are not aware of any systematic study that tested this procedure  
on a set of diverse languages and tasks.

To fill this gap,  
we rigorously evaluated the method using four typologically diverse languages (Spanish, Russian, Vietnamese, and Hindi)
and four different tasks all of which are crucial to information retrieval and 
text mining: natural-language inference (NLI), question-answering (QA), 
named-entity recognition (NER), and cross-lingual sentence retrieval (XSR).
QA and XSR are core information mining tasks directly targeting a user's information need \cite{baeza1999modern,standfordIR2008} 
while NLI and NER are used in document \cite{DBLP:phd/basesearch/Abdelrahman20} and
query \cite{DBLP:conf/bdc/WenVLWG19,DBLP:conf/aaai/ChengBBGPJ21} analysis.
Named entities are the core of entity-oriented~\cite{DBLP:series/irs/Balog18} and expert search \cite{DBLP:journals/corr/abs-2106-07742}.
Entity-oriented retrieval models can also be used to boost recall of traditional keyword-based retrievers~\cite{10.1145/3511808.3557588,EnsanB17}.

\begin{figure*}[tb]
\centering
\includegraphics[width=\linewidth]{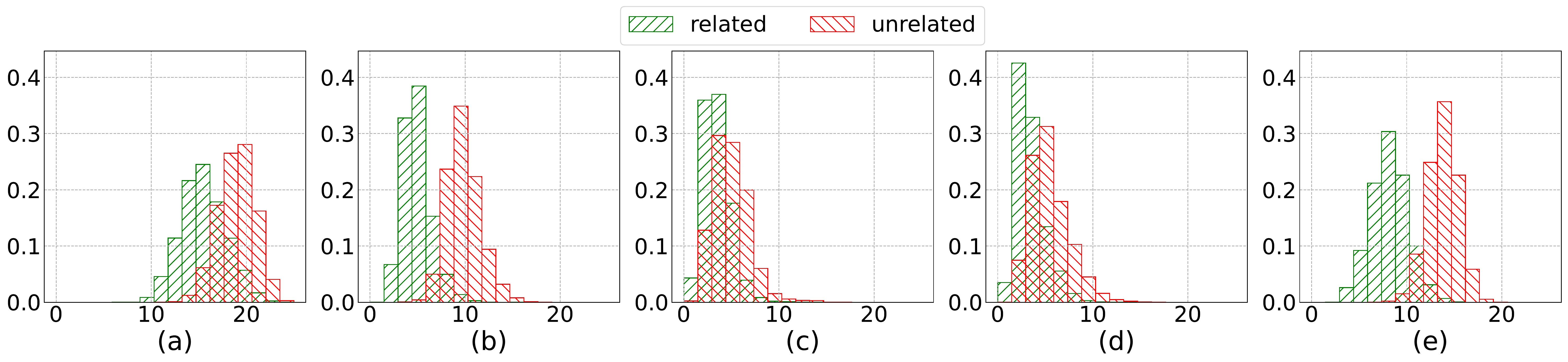}
\caption{Histograms of $L_2$ distances between pairs of mBERT 
\emph{last-layer} representations 
for randomly sampled related (i.e., aligned) and unrelated word pairs from WikiMatrix (Hi-En): (a) original\protect\footnotemark, (b)~after cross-lingual adjustment, (c)~after fine-tuning on English NLI data, (d) after cross-lingual adjustment and subsequent fine-tuning on English NLI data, (e) after cross-lingual adjustment and subsequent \emph{continual} fine-tuning on English NLI data.}
\label{fig:l2_hist}
\end{figure*}

\footnotetext{To simplify the presentation, parts of the plots in Fig.~\ref{fig:l2_hist}a corresponding to small cumulative probabilities have been cropped in the camera-ready version of the paper.}

Furthermore, existing studies typically avoided statistical testing and presented results for a single seed.
In contrast, we performed each experiment with five seeds and assessed statistical significance of the difference from a baseline.
In addition to the original method of \newcite{cao2020multilingual}
we evaluated its \emph{continual-learning} extension \cite{ratcliff1990connectionist,robins1995catastrophic,ParisiKPKW19} 
where we combined the target task loss with the cross-lingual adjustment loss \cite{Caruana96}.

In our study, we ask the following research questions:
\begin{itemize}
    \item[R1] How does cross-lingually adjusted mBERT fine-tuned on English data and zero-shot transferred to a target language perform on various IR/NLP tasks and typologically different languages?
    \item[R2] How does adjustment of mBERT on parallel data and fine-tuning for a specific task 
    affect similarity of contextualized embeddings of semantically related and unrelated words across languages?
    \item[R3] Inspired by our observation (see Fig. \ref{fig:l2_hist}c-\ref{fig:l2_hist}d) that fine-tuning 
              draws embeddings of \emph{both} related and unrelated words closer to each other,
              which may negatively affect the cross-lingual transfer,
              we wonder if continual learning---with an auxiliary cross-lingual adjustment loss---can 
              improve effectiveness of the zero-shot transfer.
\end{itemize}

Our experiments demonstrated the following:

\begin{itemize}
    \item The cross-lingual adjustment of mBERT improves NLI in four target languages; NER, XSR,
    and \emph{cross-lingual} QA in three languages (though some cross-lingual QA gains are not statistically significant).
    Yet, there is no statistically significant improvement for \emph{mono-lingual} QA 
    and a statistically significant deterioration on three out of eight QA datasets. 
    
    \item When comparing $L_2$ distances between contextualized-embeddings of words across languages (Fig.~\ref{fig:l2_hist}b),
    we  see that the cross-lingual adjustment of mBERT decreases the $L_2$ distance between related words while keeping unrelated words apart,
    which is in line with prior work \cite{zhao2021inducing}.
    
    \item However, we have found no prior work that inspected histograms obtained after fine-tuning.
    Quite surprisingly, we observe that fine-tuning of mBERT for a specific task 
    draws embeddings of \emph{both} related and unrelated words much closer to each other (Fig.~\ref{fig:l2_hist}c and Fig.~\ref{fig:l2_hist}d).
    Thus, fine-tuning causes the model to ``forget'' some of the cross-lingual information learned during adjustment.
            
    \item In that, continual learning allows the model to learn a target task while maintaining the separation between related and
    unrelated words (Fig.~\ref{fig:l2_hist}e). 
    Continual learning consistently improves performance on NLI data.
    Aside from NLI, it improves XSR and NER \emph{only} for Russian and there are no improvements on other QA or NER datasets.
    
\end{itemize}

In summary, our study contributes to a better understanding of (1)~cross-lingual transfer capabilities of large multilingual language models and of (2)~effectiveness of their cross-lingual adjustment across various tasks and languages.
We believe our results support a conjecture that the cross-lingual adjustment of \newcite{cao2020multilingual} is more beneficial for cross-lingual tasks.
Our software is publicly available \url{https://github.com/pefimov/cross-lingual-adjustment}.

\section{Related Work} \label{sec:RW}

\subsection{Cross-Lingual Zero-Shot Transfer with Multilingual Models}

The success of mBERT in cross-language zero-shot regime on various tasks has inspired many papers 
that attempted to explain its cross-lingual abilities and limitations~\cite{wu2019beto,wu2019emerging,k2020cross,libovicky2019language,dufter2020identifying,chi2020finding,pires2019multilingual,xquad,chi2020finding}. 
These studies showed that the multilingual models learn high-level abstractions common to all languages, which make transfer possible even when languages share no vocabulary.
However, the gap between performance on English and a target language is smaller if the languages are cognate, i.e. share a substantial portion of model's vocabulary, have similar syntactic structures, and are from the same language family~\cite{wu2019beto,lauscher2020zero}. Moreover, the size of target language data used for pre-training and the size of the model vocabulary allocated to the language also positively impacts cross-lingual learning performance~\cite{lauscher2020zero,xquad}.

Zero-shot transfer of mBERT or other multilingual transformer-based models from English to a different language was applied to many tasks including but not limited 
to cross-lingual information retrieval, POS tagging, dependency parsing, NER, NLI, and QA~\cite{wu2019beto,wang2019cross,pires2019multilingual,hsu2019zero,litschko2021evaluating}. 
XTREME data suite~\cite{hu2020xtreme} and its successor \mbox{XTREME-R}~\cite{ruder2021xtreme} are dedicated collections of tasks and corresponding datasets for evaluation of zero-shot transfer capabilities of large multilingual models from English to tens of languages. 
XTREME includes XSR, NLI, NER, and QA datsets used in the current study.
Although transfer from English is not always an optimal choice~\cite{lin2019choosing,turc2021revisiting},
English still remains the most popular source language. 
Furthermore, despite there have been developed
quite a few new models that differ in architectures, supported languages, and training data  ~\cite{doddapaneni2021primer}, mBERT remains the most popular cross-lingual model.

\subsection{Cross-lingual Alignment of Embeddings}

\newcite{mikolov2013exploiting} demonstrated that vector spaces can encode semantic relationships between words
and that there are similarities in the geometry of these vectors spaces
across languages.
A variety of approaches have been proposed for aligning monolingual representations based on bilingual dictionaries and parallel sentences. 
The most widely used approach---which requires only a bilingual dictionary---consists in finding a rotation matrix that aligns vectors of two monolingual models ~\cite{mikolov2013exploiting}. 
\newcite{lample2018word} proposed an alignment method based on adversarial training, which does not require parallel data. 
\newcite{ruder2019survey} provide a comprehensive overview of alignment methods for pre-Transformer models.

\newcite{schuster2019cross} applied rotation  to align contextualized ELMo embeddings~\cite{elmo} using ``anchors'' (averaged vectors of tokens in different contexts) and bilingual dictionaries. 
They showed improved results of cross-lingual dependency parsing using English as source and several European languages as target languages. \newcite{wang2019parsing} aligned English BERT and mBERT representations using rotation method and Europarl parallel data~\cite{koehn2005europarl}. They employed the resulting embeddings in a cross-lingual dependency parsing model. 
The parser with aligned embeddings consistently outperformed zero-shot mBERT on 15 out of 17 target languages.
Instead of aligning on a word level, \newcite{aldarmaki2019context} 
performed a sentence-level alignment of ELMo embeddings and evaluated 
this approach on the parallel sentence retrieval task.

\newcite{cao2020multilingual} proposed to directly modify the mBERT model 
by bringing the vectors of semantically related words in different languages closer to each other.  
This was motivated by the observation that embedding spaces of different languages are not 
always isometric~\cite{sogaard2018limitations} and, hence, are not always amenable to alignment via rotation. 
The authors showed that mBERT simultaneously adjusted on five European languages consistently outperformed other alignment approaches on XNLI data.
%~\cite{conneau2018xnli}. 
In the current study, we implement the approach with some modifications.

\newcite{LiuWMF21} showed that combining continual learning with fine-tuning improved zero-shot transfer performance for NER and POS tagging.
In that, they used cross-lingual sentence retrieval (XSR) and/or masked-language model (MLM) task as additional tasks.
Although XSR can be seen as an alternative to the cross-lingual adjustment of \newcite{cao2020multilingual},
the authors did not evaluate the effectiveness of zero-shot transfer after adjusting the model with XSR.
In contrast, we evaluate the marginal effectiveness of continual learning with respect to already cross-lingually adjusted mBERT.

\newcite{kulshreshtha2020cross} compared different alignment methods (rotation vs. adjustment) 
on NER and slot filling tasks. 
According to their results, rotation-based alignment performs better on the NER task, while model adjustment performs better on slot filling. 
\newcite{zhao2021inducing} continued this line of research and proposed several improvements of the model adjustment method:
1)~z-normalization of vectors
and 2)~text normalization to make the input more structurally `similar' to English training data.
Experiments on XNLI dataset and translated sentence retrieval showed that vector normalization leads to more consistent improvements over zero-shot baseline compared to text normalization. 
\newcite{faisal-anastasopoulos-2021-investigating} applied cross-lingually adjusted mBERT and XLM-R to cross-lingual open-domain QA and obtained improvements both on paragraph and span selection subtasks. However, they trained their models on machine-translated data, which is different from our zero-shot settings.

\section{Methods}

In this study, we use a multilingual BERT (mBERT) as the main model \cite{bert}.
mBERT is a case-sensitive ``base'' 12-layer Transformer model~\cite{VaswaniSPUJGKP17} with 178M parameters.\footnote{\url{https://huggingface.co/bert-base-multilingual-cased}} 
It was trained with a masked language model objective on 104 languages with a shared WordPiece \cite{WuSCLNMKCGMKSJL16} vocabulary (using 104 Wikipedias).
To balance the distribution of languages, high-resource languages were under-sampled and low-resource languages were over-sampled.\footnote{\url{https://github.com/google-research/bert/blob/master/multilingual.md}} 
For a number of NLP tasks,
cross-lingual transfer of mBERT can be competitive with training a monolingual model using the training
data in the target language.

We align cross-lingual embeddings by directly modifying/adjusting the language model itself, 
following the approach by \newcite{cao2020multilingual}. The approach---which differs from finding a rotation matrix---proved to be effective in the NLI task.
However, there are some differences in our implementation. 
In all cases, we work with one pair of languages at a time 
while \newcite{cao2020multilingual} adjusted mBERT for five languages at once.

BERT uses WordPiece tokenization \cite{WuSCLNMKCGMKSJL16}, 
which splits sufficiently long words into \emph{subword} tokens. 
We first word-align parallel data with \textit{fast\_align}~\cite{fastalign} and then average all subword tokens' vectors.\footnote{We also experimented with other options reported in the literature -- first/last tokens' vectors, as well as aligning subword tokens produced by BERT. Although these choices induced some variations in results, there is no single pattern across all tasks and languages.}

Based on alignments in parallel data, 
we obtain a collection of word
pairs $(s_i, t_i)$: $s_i$ from the source language, $t_i$ from the target one.
From these alignments we can obtain their mBERT vector representations $\mathbf{f}(s_i)$ and $\mathbf{f}(t_i)$.
Then, we \emph{adjust} the mBERT model
on aligned pairs' vectors using the following loss function:
\begin{equation} \label{eq:cross_ling_adjust_loss}
L= \sum_{(s_i, t_i)} \|\mathbf{f}(s_i)  - \mathbf{f}(t_i) \|_2^2 + \sum_{s_j}  \| \mathbf{f}(s_j) - \mathbf{f}^0(s_j) \|_2^2,
\end{equation}
\noindent where the first term ``pulls'' the embeddings in the source and target language together, 
while the second (regularization) term prevents source (English) representations from deviating far from their initial values in the `original' mBERT $\mathbf{f}^0$.
Finally, the cross-lingually adjusted mBERT model is fine-tuned for a specific task. 

Training neural networks via empirical loss minimization is known to suffer from the ``catastrophic forgetting'' \cite{mccloskey1989catastrophic}.
The histograms of $L_2$ distances between embeddings of related and unrelated words in pairs of languages (see Fig. ~\ref{fig:l2_hist} and
the discussion in \S~\ref{sec:bert_analysis}), confirm that this is, indeed, the case.
Specifically, fine-tuning on a target task---in contrast to the cross-lingual adjustment objective---reduces the separation between related and unrelated words.
To counter this effect, we ran an additional experiments in a continual-learning  mode \cite{riabi2020synthetic},
which relies on experience replay \cite{ratcliff1990connectionist,robins1995catastrophic}.

Technically, this entailed a multi-task  training \cite{Caruana96} with a combined loss function:
\begin{equation} \label{eq:continual_loss}
 L = L_{target} + \alpha L_{align},
\end{equation}
where $L_{target}$ was the loss-function for the target task, e.g., 
NLI, $L_{align}$ was a cross-lingual loss function given by Eq.~\ref{eq:cross_ling_adjust_loss}, and $\alpha>0$ was a small weight.
During training, we iterated over the complete (reshuffled) dataset for the target task: 
After computing $L_{target}$ for a current batch we randomly sampled a small batch of aligned pairs of words $\{ (s_i, t_i) \}$ from the parallel corpus and computed $L_{align}$.

\section{Tasks and Data}
\subsection{Languages and Parallel Data} \label{sec:parallel_data}

\begin{wraptable}{r}{0.50\linewidth}
{
\setlength{\tabcolsep}{5pt}
    \centering
\begin{threeparttable}
    \scriptsize
    \begin{tabular}{lllcc}
      \hline
          Lang  & Family & Script & Word  & Number of English              \\ 
                &        &        & order &   Wiki pages                   \\  \hline
         en     & IE/Germanic & Latin & SVO &  6.3M \\
         es     & IE/Romance & Latin & SVO & 1.7M \\
         ru      & IE/Slavic & Cyrillic & SVO & 1.7M\\
         vi     & Austroasiatic & Latin & SVO & 1.3M\\
         hi     & IE/Indo-Aryan & Devanagari & SOV & 150K\\ \hline
    \end{tabular}
    \begin{tablenotes}
        \scriptsize
        \item IE : Indo-European; Prevalent word order: SVO~-- subject-verb-object, SOV~-- subject-object-verb;
    \end{tablenotes}
        \caption{Language information.}
    \label{tab:languages}
\end{threeparttable}
}
\end{wraptable}

In our experiments we transfer models trained on English to four languages: Spanish, Russian, Vietnamese, and Hindi. This set represents four different families (including one non-Indo-European language), three scripts,
and two different prevalent word orders (see Table~\ref{tab:languages}). 
All the languages are among languages that were used to train mBERT.\footnote{However, Hindi Wikipedia
is an order of magnitude smaller compared to other Wikipedias, which may have led to somewhat inferior contextualized embeddings.}

To align embeddings, we use a sample from the parallel corpus WikiMatrix~\cite{wikimatrix} that contains sentences in 1,620 different language pairs mined from Wikipedia.

\subsection{Natural Language Inference}\label{sec:XNLI}
Natural language inference (NLI) is a task of determining the relation between two ordered
sentences (hypothesis and premise) and classifying them into: entailment, contradiction, or ``no relation''.
English MultiNLI collection consists of 433K 
multi-genre sentence pairs~\cite{williams2017broad}. 
The XNLI dataset
complements the MultiNLI training set with newly collected 2.5K development and 5K test English examples~\cite{conneau2018xnli}.
They were professionally translated into 15 languages,
including all four target languages of the current study. 
Additionally, for each target language test set,
we created a new mixed-language XNLI  set 
by randomly picking either a hypothesis or a premise and replacing it with the original English sentence.
Performance on  XNLI datasets is evaluated using classification \emph{accuracy}.

\subsection{Named Entity Recognition}

Named entity recognition (NER) is a task of locating named entities
in unstructured text and classifying them into predefined
categories such as persons, organizations, locations, etc.
In our experiments, we employ the Wikiann NER corpus \cite{rahimi2019massively} that is derived from a larger ``silver-standard'' collection that was created fully automatically~\cite{pan2017cross}. 
Wikiann NER has data for 41 languages, 
including all languages in the current study.
The named entity types include location (LOC), person (PER), 
and organization (ORG).
The English training set contains 20K sentences. 
Test sets for Spanish, Vietnamese, and Russian have 10K sentences each; for Hindi -- 1K sentences. 
Performance is evaluated using the \emph{token-level micro-averaged F1}.

\subsection{Question Answering}

Machine reading comprehension (MRC) is a variant of a QA task. 
Given a question and a text paragraph, the system needs to return a continuous span of paragraph tokens as an answer.
The first large-scale MRC dataset is the English Wikipedia-based dataset SQuAD \cite{squad},
which contains about 100K paragraph-question-answer triples. 
SQuAD has become a \textit{de facto} standard and inspired creation of analogous resources in other languages \cite{rogers2021qa}. 
We use SQuAD as the source dataset to train MRC models. To test the models, we use XQuAD, MLQA, and TyDi QA datasets. XQuAD~\cite{xquad} is a professional translation  of 240 SQuAD paragraphs and 1,190 questions-answer pairs into 10 languages (including four languages of our study). MLQA~\cite{mlqa} data is available for six languages including Spanish, Vietnamese, and Hindi (but it does not have Russian). 
There are about 5K questions for each of our languages. 
TyDi QA~\cite{tydiqa} includes 11 typologically diverse languages of which we use only Russian (812 test items).

In addition to monolingual test data,
we experimented with parallel/cross-lingual MLQA and XQuAD datasets and explored two directions: (1)~question is in a target language, but paragraph is in English; (2)~a question is in English, but a paragraph is in a target language.

QA performance is evaluated using a \emph{token-level F1-score}.

\subsection{Cross-Lingual Sentence Retrieval}
The task of cross-lingual sentence retrieval (XSR) consists in retrieving sentences that are translations of each other.
A query is a sentence in one language and a corpus is a set of sentences (translations) in another language (in our case English).
For the XSR task, we use a subset of the Tatoeba collection~\cite{artetxe2019massively}
covering 93 languages (including four languages in our study).
This subset has one thousand English sentences each of which has a translation in each of the 93 languages.
Following \newcite{ruder2021xtreme}, we fine-tune the model on a QA task. 
A sentence representation is obtained by averaging token embeddings in one of the layers.
Representations are compared using the cosine similarity.
Because different layers perform differently in different scenarios (original mBERT vs. adjusted vs. adjusted
with continual learning), we select the scenario-specific best-performing layer.
In that, retrieval performance is measured using \emph{the mean reciprocal rank (MRR)}.

\section{Experimental Results and Analysis}

\subsection{Setup}

All experiments were conducted on a single Tesla V100 16GB. 
For cross-lingual model adjustment we use the Adam optimizer and hyper-parameters provided by \newcite{cao2020multilingual}. To obtain reliable results we run five iterations (using different seeds) of model adjustment 
(for each configuration) followed by fine-tuning on downstream tasks. 
For each run we sample 30K
sentences from 
a set of 250K parallel (WikiMatrix) sentences word-aligned with \textit{fast\_\!align}.\footnote{We ran the main experiments with 30K parallel sentences. In addition, we conducted experiments with 5K/10K/30k/100K/250K Ru-En sentence pairs.
Increasing amount of parallel data benefits both NLI and NER, whereas QA performance peaks at roughly 5K parallel sentences and further decreases as the number of parallel sentences increases. Due to limited space we don't report detailed results and analysis here.}

The code to fine-tune mBERT on XNLI, SQuAD, and Wikiann is based on  \texttt{HuggingFace} sample scripts,\footnote{\url{https://github.com/huggingface/transformers/tree/master/examples/pytorch}} which were modified to support continual learning. 
We use a standard architecture consisting of a BERT model with a task-specific linear layer \cite{bert}.
We also reuse parameters provided by \texttt{HuggingFace}, except for the weight $\alpha=0.01$ in the multi-task loss (Eq.~\ref{eq:continual_loss}),
which was tuned on XNLI validation sets. Also note that batch sizes are  32 (for the main target loss) and 16 (for the auxiliary
cross-lingual adjustment loss in the case of continual learning).

All reported results are averages over five runs with different seeds. 
We compute statistical significance of differences between the original and adjusted mBERT 
using a paired t-test. 
For XSR, QA, and NLI we first average metric values for each example over different runs 
and then carry out a paired t-test using averaged values. 
For NER we concatenate example-specific predictions for all seeds and run 1,000 iterations of a permutation test
for concatenated sequences \cite{pitman1937significance,EfronT93}.

\subsection{Main Results}

\begin{table}[tpb]
\begin{minipage}[ct]{0.50\textwidth}
\centering
\setlength{\tabcolsep}{5pt}
\begin{threeparttable}
    \small
    \begin{tabular}{lcccc}
        mBERT  &   es\phantom{**}  &  ru\phantom{**}   & vi\phantom{**} & hi\phantom{**}  \\ \hline
         \multicolumn{5}{c}{Original XNLI} \\ \hline
        Original         & 74.20\phantom{**}  &  67.95\phantom{**}  &  69.58\phantom{**}  &  59.03\phantom{**}    \\
        Adjusted         & 74.82*\phantom{*} &  69.45*\phantom{*} &  70.88*\phantom{*} &  61.54*\phantom{*}   \\
        Adj+cont & \textbf{75.89**} &  \textbf{71.26**} &  \textbf{72.79**} &  \textbf{63.90**}   \\
        \hline
        \multicolumn{5}{c}{Mixed-language NLI} \\
        \hline
        Original         & 70.93\phantom{**}  & 64.24\phantom{**}   & 62.72\phantom{**}   &  53.53\phantom{**}    \\
        Adjusted         & 72.06*\phantom{*} & 66.56*\phantom{*}  & 66.50*\phantom{*}  &  57.31*\phantom{*}   \\
        Adj+cont & \textbf{73.50**} & \textbf{69.09**}  & \textbf{69.14**}  &  \textbf{61.09**}   \\
         \hline
    \end{tabular}
    \begin{tablenotes}
        \item Statistically significant differences from an original 
        and adjusted mBERT are marked with * and **, respectively (p-value threshold 0.05).
    \end{tablenotes}
\caption{Performance on original and mixed-language NLI datasets (accuracy).}
\label{tab:xnli_results}
\end{threeparttable}
\end{minipage}
\hspace{\fill}
\begin{minipage}[ct]{0.50\textwidth}
\centering
\setlength{\tabcolsep}{5pt}
\begin{threeparttable}
    \small
    \begin{tabular}{lcccc}
        mBERT            & es\phantom{**}     &   ru\phantom{**}    & vi\phantom{**}      & hi\phantom{**}        \\ \hline
        Original         & \textbf{73.40}\phantom{**}  &  63.43\phantom{**}  & 71.02\phantom{**}   & 65.24\phantom{**}     \\
        Adjusted         & 73.28\phantom{**}  &  65.49*\phantom{*}  & \textbf{71.99*}\phantom{*}   & \textbf{68.22*}\phantom{*}     \\
        Adj+cont & 72.71**  &  \textbf{66.27**}  & 71.35**   & 66.07**     \\
         \hline
    \end{tabular}
    \begin{tablenotes}
        \item Statistically significant differences from an original 
        and adjusted mBERT are marked with * and **, respectively (p-value threshold 0.05).
    \end{tablenotes}
\caption{Performance on NER task (token-level F1).}
\label{tab:ner_results}
\end{threeparttable}
\end{minipage}
\end{table}

\begin{table*}[tb]
\centering
\begin{threeparttable}
     \small
    \begin{tabular}{c|cc|cc|cc|cc}
    \hline
    \multirow{2}{*}{mBERT}  & \multicolumn{2}{c|}{Spanish} & \multicolumn{2}{c|}{Russian} & \multicolumn{2}{c|}{Vietnamese} & \multicolumn{2}{c}{Hindi}\\   
                     &   \multicolumn{1}{c}{MLQA}  &   \multicolumn{1}{c|}{XQuAD}   &  \multicolumn{1}{c}{TyDi QA} &   \multicolumn{1}{c|}{XQuAD}   &   \multicolumn{1}{c}{MLQA}  &  \multicolumn{1}{c|}{XQuAD}   &  \multicolumn{1}{c}{MLQA}   &  \multicolumn{1}{c}{XQuAD}   \\ \hline
    %                     |       Spanish |   Russian     |   Vietnamese  |     Hindi     |
    %                     |   ML  |   XQ  |  TyDi |   XQ  |   ML  |  XQ   |  ML   |  XQ   |
         Original         & \textbf{64.96}\phantom{**} & \textbf{75.59}\phantom{**} & 67.05\phantom{**} & \textbf{70.72}\phantom{**} & \textbf{59.95}\phantom{**} & \textbf{69.18}\phantom{**} & \textbf{48.73}\phantom{**} & 57.56\phantom{**}  \\
         Adjusted         & 63.11*\phantom{*}& 73.99*\phantom{*}& 67.03\phantom{**} & 70.58\phantom{**} & 58.46*\phantom{*}& 68.63\phantom{**} & 48.47\phantom{**} & 57.81\phantom{**}  \\
         Adj+cont & 62.76**& 73.44\phantom{**} & \textbf{67.63}\phantom{**} & 70.51\phantom{**} & 57.71**& 68.64\phantom{**} & 48.02**& \textbf{57.83}\phantom{**}  \\ 
         \hline
                          \multicolumn{9}{c}{Question in target language, paragraph in English}     \\ \hline
         Original         & \textbf{67.34}\phantom{**} & \textbf{75.74}\phantom{**} &  --   & 71.54\phantom{**} & 56.08\phantom{**} & 65.00\phantom{**} & 42.48\phantom{**} & 47.83\phantom{**}  \\
         Adjusted         & 66.93*\phantom{*}& 75.65\phantom{**} &  --   & \textbf{71.68}\phantom{**} & \textbf{56.74*}\phantom{*}& \textbf{66.75*}\phantom{*}& \textbf{44.91*}\phantom{*}& \textbf{50.45*}\phantom{*} \\
         Adj+cont & 66.31**& 74.88**&  --   & 70.99\phantom{**} & 54.51**& 64.63**& 43.88**& 50.13\phantom{**}  \\
         \hline
                         \multicolumn{9}{c}{Question in English, paragraph in target language}    \\ \hline
         Original         & \textbf{67.36}\phantom{**} & \textbf{76.71}\phantom{**} &  --   & 67.31\phantom{**} & 64.43\phantom{**} & 68.12\phantom{**} & 55.32\phantom{**} & 58.62\phantom{**}  \\
         Adjusted         & 66.96*\phantom{*}& 76.42\phantom{**} &  --   & \textbf{68.25*}\phantom{*}& \textbf{65.01*}\phantom{*}& \textbf{68.99}\phantom{**} & \textbf{55.63}\phantom{**} & \textbf{58.93}\phantom{**}  \\
         Adj+cont & 66.68**& 76.21\phantom{**} &  --   & 68.06\phantom{**} & 64.36**& 68.54\phantom{**} & 54.74**& 58.22\phantom{**}  \\
         \hline
         
    \end{tabular} 
    \begin{tablenotes}
        \item Statistically significant differences from an original 
        and adjusted mBERT are marked with * and **, respectively (p-value threshold 0.05).
    \end{tablenotes}
\caption{Effectiveness of QA systems (F1-score).} \label{tab:qa_results}
\end{threeparttable}
\end{table*}

Results for NLI, NER, QA, and XSR tasks are summarized in Tables~\ref{tab:xnli_results}, \ref{tab:ner_results}, \ref{tab:qa_results}, and \ref{tab:tatoeba_results}, respectively. 
We can observe consistent and  statistically significant improvements (up to 2.5 accuracy point) of aligned models over zero-shot transfer on XNLI for all languages.
This is in line with \newcite{cao2020multilingual} 
even though we used a set of more diverse languages, parallel data of lower quality, 
and a slightly different learning scheme (we adjusted models individually for each pair of languages).
In that, employing continual learning lead to additional substantial gains (up to 2.4 accuracy points).

We also evaluated models on the (bilingual) mixed-language XNLI test data (see \S~\ref{sec:XNLI}).
According to the bottom part of Table~\ref{tab:xnli_results},
compared to the original XNLI,
we observe bigger gains for all four languages,
especially when we employ continual learning.
For Hindi, we obtain a 7.5 point gain by using both the adjustment and continual learning.

NER results are somewhat mixed: 
We observe statistically significant gains (up to 3 points for Hindi) on all languages except Spanish. 
In that, continual learning is beneficial only for Russian.

When we fine-tuned a cross-lingually adjusted mBERT on \emph{mono-lingual} QA tasks,
there were no statistically significant gains.
In that, there was a statistically significant decrease for all Spanish datasets and Vietnamese MLQA.
Use of continual learning lead to further degradation in nearly all cases.
Note that models were noticeably less accurate on MLQA compared to XQuAD,
which is a translation of our training set SQuAD.

\newcite{muttenthaler2020unsupervised} and \newcite{vanaken2019does} showed that 
QA models essentially clustered answer token vectors and separated them from the rest of the paragraph token vectors using a vector representation of the question.
Thus, to solve the QA task, 
the model learns to rely on \emph{mutual similarities} among question and answer tokens
(on English QA data) rather than on their actual vector representations.
As a consequence, 
there may be no need to make representations in the target language to be similar to English-language representations.
This may \emph{partially} explain why the cross-lingual adjustment was unsuccessful for \emph{mono-lingual} QA.

This hypothesis---together with the observation of stronger performance on XQuAD compared to MLQA---prompted us to explore whether the cross-lingual adjustment could be more useful for \emph{cross-lingual} QA.
We explored two directions: (1)~question is in a target language, but paragraph is in English; (2)~a question is in English, but a paragraph is in a target language. 
According to results in the lower part of Table~\ref{tab:qa_results},
we observe improvements in three languages (except Spanish) and most of these improvements are statistically significant.
This is in line with cross-lingual QA results by \newcite{faisal-anastasopoulos-2021-investigating}. 

\begin{wraptable}{r}{0.5\textwidth}
\centering
\begin{threeparttable}
    \small
    \begin{tabular}{lcccc}
        mBERT    &  es\phantom{**}  &  ru\phantom{**}   & vi\phantom{**}   & hi\phantom{**}   \\ \hline
        Original & 0.80\phantom{**} & 0.73\phantom{**}	& 0.67\phantom{**} & 0.49\phantom{**} \\
        Adjusted & \textbf{0.82*}\phantom{*} & 0.72*\phantom{*}  & \textbf{0.72*}\phantom{*} & \textbf{0.56*}\phantom{*} \\
        Adj+cont & 0.78** & \textbf{0.74**}  & 0.69** & 0.53** \\
         \hline
    \end{tabular}
    \centering
    \begin{tablenotes}
        \item Statistically significant differences from an original 
         and adjusted mBERT are marked with * and **, respectively (p-value threshold 0.05).
     \end{tablenotes}
\caption{\label{tab:tatoeba_results}Performance on XSR task (MRR for best-performing layers).}
\end{threeparttable}
\end{wraptable}

For XSR---another cross-lingual task---using the cross-lingual adjustment lead to substantial and statistically significant improvements in three languages: Spanish, Vietnamese, and Hindi. However, continual learning only marginally helped Russian.
Finally, we note that cross-lingual adjustment resulted in bigger gains for the mixed,
i.e., cross-lingual XNLI data, than for original one (Table~\ref{tab:xnli_results}).
In summary, 
we believe our results support a conjecture that the cross-lingual adjustment of \newcite{cao2020multilingual} is more beneficial for cross-lingual tasks.

\subsection{Analysis of the Adjusted mBERT for NLI}\label{sec:bert_analysis}
This section analysis focuses on the NLI task.
We calculate $L_2$ distances between contextualized embeddings in English and other languages.\footnote{Although most prior work uses the cosine similarity instead of $L_2$ \cite{RudmanGRE22},
it does not distinguish between vectors with the same direction, but different lengths.
}
The embeddings are taken from the last layer output (i.e.,
no prediction heads are used).
To this end we sampled 
semantically related words from parallel sentences (matched via \emph{fast\_\!align})
and unrelated words from unpaired sentences (nearly always unrelated).
For each pair of languages and each task, the sampling process was carried out 
for: (1) the original mBERT, (2) an adjusted mBERT, (3) the original mBERT fine-tuned
for the target task, (4) the adjusted mBERT fine-tuned for the target task,
(5) the adjusted mBERT fine-tuned for the target task using \emph{continual} learning.\footnote{Due to limited space we demonstrate histograms only for Hindi/NLI task, other languages/tasks exhibit similar trends.}

From Fig.~\ref{fig:l2_hist}
we can see that the cross-lingual adjustment makes embeddings of semantically similar words from different languages closer to each other 
while keeping unrelated words apart, which is in line with \newcite{zhao2021inducing}.
However, prior work did not inspect histograms obtained after fine-tuning.
Yet, quite surprisingly, fine-tuning of both the original and adjusted mBERT on the English NLI data  
(Fig.~\ref{fig:l2_hist}c  and \ref{fig:l2_hist}d) 
makes distributions of related and unrelated words almost fully overlap, 
i.e. all embeddings become close to each other.
Compared to the original mBERT,
fine-tuning of the \emph{adjusted} mBERT (Fig. \ref{fig:l2_hist}d) does result in a better separation
of related and unrelated words, but the effect is \emph{quite modest}.
We believe this is an example of ``catastrophic forgetting`` \cite{mccloskey1989catastrophic}, where
fine-tuning the model on a target task causes the model to forget some of the knowledge obtained during cross-lingual adjustment.

Continual learning (Fig.~\ref{fig:l2_hist}e) permits fine-tuning for the target task while maintaining a separation between related and unrelated words, which also consistently improves performance for the NLI task.
However, when  we compared histograms for additional tasks (not shown due to space limit)  there was no direct relationship between 
the degree of separation and the success of cross-lingual transfer among all tasks and training regimes.
In the case of NER, the biggest separation was achieved for Spanish,
but fine-tuning of the adjusted mBERT resulted in a lower accuracy.
More generally, fine-tuning with continual learning \emph{always} led to a better separation of related and unrelated words, but this extra separation was beneficial only for the NLI task.

\section{Conclusion}

We evaluate effectiveness of an existing approach to cross-lingual adjustment of mBERT \cite{cao2020multilingual}
using  four typologically different languages (Spanish, Russian, Vietnamese, and Hindi) and four IR/NLP tasks (XSR, QA, NLI, and NER).
The original mBERT is being compared to mBERT ``adjusted'' with a help of a small parallel corpus.
The cross-lingual adjustment of mBERT improves NLI in four target languages; NER, XSR,
    and \emph{cross-lingual} QA in three languages (though some cross-lingual QA gains are not statistically significant).
However, in the case of \emph{mono-lingual }QA performance never improves and sometimes degrades.
We believe our results support a conjecture that the cross-lingual adjustment of \newcite{cao2020multilingual} is more beneficial for cross-lingual tasks.

Inspired by the analysis of histograms of distances,
we obtain \emph{additional} improvement on NLI using continual learning.
Our study contributes to a better understanding of cross-lingual transfer capabilities of large multilingual language models and  identifies limitations of the cross-lingual adjustment in various IR and NLP tasks.

\section*{Acknowledgments} This research was supported in part through computational resources of HPC facilities at HSE University \cite{hpc-hse}. PE is grateful to Yandex Cloud for their grant toward computing resources of Yandex DataSphere.\footnote{\url{https://cloud.yandex.com/en/services/datasphere}} PB acknowledges support by the Russian Science Foundation, grant № 20-11-20166.

\bibliographystyle{coling}
\bibliography{coling2020}

\end{document}